\pdfoutput=1

\documentclass[11pt]{article}

\usepackage[]{emnlp2021}
\usepackage{url}            
\usepackage{booktabs}       
\usepackage{amsfonts}       
\usepackage{nicefrac}       
\usepackage{microtype}      
\usepackage{subcaption}
\usepackage{amsmath,amsfonts,amssymb}
\usepackage{breqn}
\usepackage{tabularx}
\usepackage{multirow}
\usepackage{graphicx}
\usepackage{color}
\usepackage{tcolorbox}
\usepackage{CJK}
\usepackage{adjustbox}
\usepackage{xcolor}
\usepackage{colortbl}
\usepackage{multicol}
\usepackage{vwcol} 
\newsavebox{\algleft}
\newsavebox{\algright}
\usepackage{bm}
\usepackage{booktabs}
\usepackage{ctable} 
\usepackage{amsmath,stackengine}
\usepackage[linesnumbered,ruled,vlined]{algorithm2e}
\usepackage{times}
\usepackage{latexsym}
\usepackage{dsfont}
\usepackage[T1]{fontenc}

\usepackage[utf8]{inputenc}

\usepackage{microtype}

\title{CLASSIC: Continual and Contrastive Learning of Aspect \\ Sentiment Classification Tasks}

\author{
Zixuan Ke$^{1}$, Bing Liu$^{1}$, Hu Xu$^{2}$ \and Lei Shu$^{3}$\thanks{\hspace{5pt} Work was done prior to joining Amazon.} \\ 
$^1$Department of Computer Science, University of Illinois at Chicago\\
$^2$Facebook AI Research\\
$^3$Amazon AWS AI\\
$^1$\texttt{\{zke4,liub\}@uic.edu}\\  $^2$\texttt{huxu@fb.com} \\ $^3$\texttt{shulindt@gmail.com}}

\begin{document}
\maketitle
\begin{abstract}
This paper studies continual learning (CL) of a sequence of aspect sentiment classification (ASC) tasks in a particular CL setting called \textit{domain incremental learning} (DIL).~Each task is from a different domain or product.~The DIL setting is particularly suited to ASC because in testing the system needs not know the task/domain to which the test data belongs.~To our knowledge, this setting has not been studied before for ASC.~This paper proposes a novel model called CLASSIC.~The key novelty is a \textit{contrastive continual learning} method that enables both knowledge transfer across tasks and knowledge distillation from old tasks to the new task, which eliminates the need for task ids in testing. Experimental results show the high effectiveness of CLASSIC.\footnote{\url{https://github.com/ZixuanKe/PyContinual}}

\end{abstract}

\section{Introduction}
\label{sec.intro}

Continual learning (CL) learns a sequence of tasks incrementally.~After learning a task, its training data is often discarded~\cite{chen2018lifelong}. The CL setting is useful when the data privacy is a concern, i.e., the data owners do not want their data used by {\color{black} others~\cite{ke2020continual,qin2020using,ke2021adapting}.} 
In such cases, if we want to leverage the knowledge learned in the past to improve the new task learning, CL is appropriate as it shares only the learned model, but not the data. In our case, a task is a separate \textit{aspect sentiment classification} (ASC) problem of a product or domain (e.g., camera or phone)~\cite{liu2012sentiment}. 
ASC is stated as follows: Given an aspect term (e.g., \textit{sound quality} in a phone review) and a sentence containing the aspect (e.g., "\textit{The sound quality is poor}"), ASC classifies whether the sentence expresses a positive, negative, or neutral opinion about the aspect. 

There are three CL settings~\cite{Ven2019Three}: \textit{Class Incremental Learning} (CIL),~\textit{Task Incremental Learning} (TIL), and \textit{Domain Incremental Learning} (DIL). 
In CIL, the tasks contain non-overlapping classes. Only one model is built for all classes seen so far. In testing, no task information is provided. This setting is not suitable for ASC as ASC tasks have the same three classes. 
{\color{black}TIL builds one model for each task in a shared network. 
In testing, the system needs the task (e.g., phone domain) that each test instance (e.g., "\textit{The sound quality is great}") belongs to and uses only the model for the task to classify the instance. Requiring the task information (e.g., phone domain) is a limitation. Ideally, the user should not have to provide this information for a test sentence. 
That is the DIL setting, i.e., all tasks sharing the same fixed classes (e.g., positive, negative, and neutral). In testing, no task information is required.}

This work uses the DIL setting to learn a sequence of ASC tasks in a neural network. The key objective is to transfer knowledge across tasks to improve classification compared to learning each task separately. An important goal of any CL is to overcome \textit{catastrophic forgetting} (CF)~\cite{mccloskey1989catastrophic}, which means that in learning a new task, the system may change the parameters learned for previous tasks and cause their performance to degrade. We solve the CF problem as well; otherwise we cannot achieve improved accuracy. {\color{black}However, sharing the classification head for all tasks in DIL makes cross-task interfere/update inevitable. Without task information provided in testing makes DIL even more challenging.} 
Previous research has shown that one of the most effective approaches for ASC~\cite{DBLP:conf/naacl/XuLSY19,sun-etal-2019-utilizing} is to fine-tune the BERT~\cite{DBLP:conf/naacl/DevlinCLT19} using the training data. However, our experiments show that this works poorly for DIL because the fine-tuned BERT on a task captures highly task specific features that are hard to use 
by other tasks. 

In this paper, we propose a novel model called CLASSIC (\textbf{C}ontinual and contrastive \textbf{L}earning for \textbf{AS}pect \textbf{S}ent\textbf{I}ment \textbf{C}lassification) in the DIL setting. Instead of fine-tuning BERT for each task, which causes serious CF, CLASSIC uses the idea of Adapter-BERT in~\cite{DBLP:conf/icml/HoulsbyGJMLGAG19} to avoid changing BERT parameters and yet achieve equally good results as BERT fine-tuning. A novel \textit{contrative continual learning} method is proposed {\color{black}(1) to transfer the shareable knowledge across tasks to improve the accuracy of all tasks, and (2) to distill the knowledge (both shareable and not shareable) from previous tasks to the model of the new task so that \textit{the new/last task model can perform all tasks}, which eliminates the need for task information (e.g., task id) in testing.} 
Existing contrastive learning~\cite{chen2020simple} cannot do these. 

Task masks are also learned and used to protect task-specific knowledge to avoid forgetting (CF). 
Extensive experiments have been conducted to show the effectiveness of CLASSIC. 

In summary, this paper makes the following contributions:
\textbf{(1)} It proposes the problem of domain continual learning for ASC, which has not been attempted before. 
\textbf{(2)} It proposes a new model called CLASSIC that uses adapters to incorporate the pre-trained BERT into the ASC continual learning, a novel \textit{contrastive continual learning} method for knowledge transfer and distillation, and task masks to isolate task-specific knowledge to avoid CF.  

\section{Related Work}
\label{sec.related.work}


Several researchers have studied lifelong or continual learning for sentiment analysis. Early works are done under \textit{Lifelong Learning} (LL)~\cite{Silver2013,ruvolo2013ella,chen2014topic}.  
Two Naive Bayes (NB) approaches were proposed to improve the new task learning~\cite{DBLP:conf/acl/ChenM015,hao2019forward}.  
\citet{xia2017distantly} proposed a voting based approach. 
All these systems work on document sentiment classification (DSC). 
\citet{ShuXuLiu2017} used LL for aspect extraction. These works do not use neural networks, and have no CF problem. 

L2PG~\cite{qin2020using} uses a neural network but improves only the new task learning for DSC.  \citet{shuai2018lifelong} worked on ASC, but since they improve only the new task learning, they did not deal with CF. Each task uses a separate network. 

Existing CL systems SRK~\cite{DBLP:conf/dasfaa/LvWLCZ19} and KAN \cite{ke2020continual} are for DSC in the TIL setting, not for ASC. {\color{black}B-CL~\cite{ke2021adapting} is the first CL system for ASC. It also uses the idea of Adapter-BERT in~\cite{DBLP:conf/icml/HoulsbyGJMLGAG19} and is based on Capsule Network. More importantly, B-CL works in the TIL setting. The proposed CLASSIC system is based on contrastive learning and works in the DIL setting for ASC, which is a more realistic setting for practical applications. 
}

\vspace{+2mm}
\noindent
\textbf{General Continual Learning (CL):} CL has been studied extensively in machine learning~\cite{chen2018lifelong,Parisi2019continual}. 
Existing work mainly focuses on dealing with CF. There are several main approaches. (1) 
\textit{Regularization-based approaches} such as those in~\cite{Kirkpatrick2017overcoming,DBLP:conf/nips/LeeKJHZ17} add a regularization in the loss to consolidate previous knowledge when learning a new task. (2) 
\textit{Parameter isolation-based approaches} such as those in \cite{Serra2018overcoming,ke2020mixed,abati2020conditional} make different subsets of the model parameters dedicated to different tasks and identify  
and mask them out during the training of the new task.
(3) \textit{Replay-based approaches} such as those in~\cite{Rebuffi2017,Lopez2017gradient,Chaudhry2019ICLR} retain an exemplar set of old task training data to help train the new task. The methods in~\cite{Shin2017continual,Kamra2017deep,Rostami2019ijcai,He2018overcoming} build data generators for previous tasks so that in learning the new task, the generated data for previous tasks can help avoid CF. 

These methods are for overcoming CF in the CIL or TIL setting of CL. Limited work has been done on knowledge transfer, which is our goal. There is little work in the DIL setting except {\color{black}the replay method DER++~\cite{buzzega2020dark}, which saves some past data. CLASSIC saves no past data.} 

Contrastive learning~\cite{chen2020simple,he2020momentum} is the base of our contrastive continual learning method. However, there is a major difference. Existing contrastive learning uses various transformations (e.g., rotation and cropping) of the existing data (e.g., images) to generate different views of the data. However, we use the hidden space information from the previous task models to create views for explicit knowledge transfer and distillation. Existing contrastive learning cannot do that. 



\section{Proposed CLASSIC Method}
\label{sec.preliminary}

State-of-the-art ASC systems all use BERT \cite{DBLP:conf/naacl/DevlinCLT19} or other language models as the base.
The proposed technique CLASSIC adopts the BERT-based ASC formulation in \cite{DBLP:conf/naacl/XuLSY19}, where the aspect term (e.g., \textit{sound quality}) and review sentence (e.g., "\textit{The sound quality is great}") are concatenated via \texttt{[SEP]}. The sentiment polarity is predicted on top of the \texttt{[CLS]} token. As indicated earlier, although BERT can achieve state-of-the-art performance on a single task, its architecture and fine-tuning are unsuitable for CL (see Sec.~\ref{sec.intro}) and perform very poorly (Sec.~\ref{sec:results}).~We found that the BERT adapter idea in~\cite{DBLP:conf/icml/HoulsbyGJMLGAG19} is a better fit for CL.  

\begin{figure}[t]
\centering
\includegraphics[width=\columnwidth]{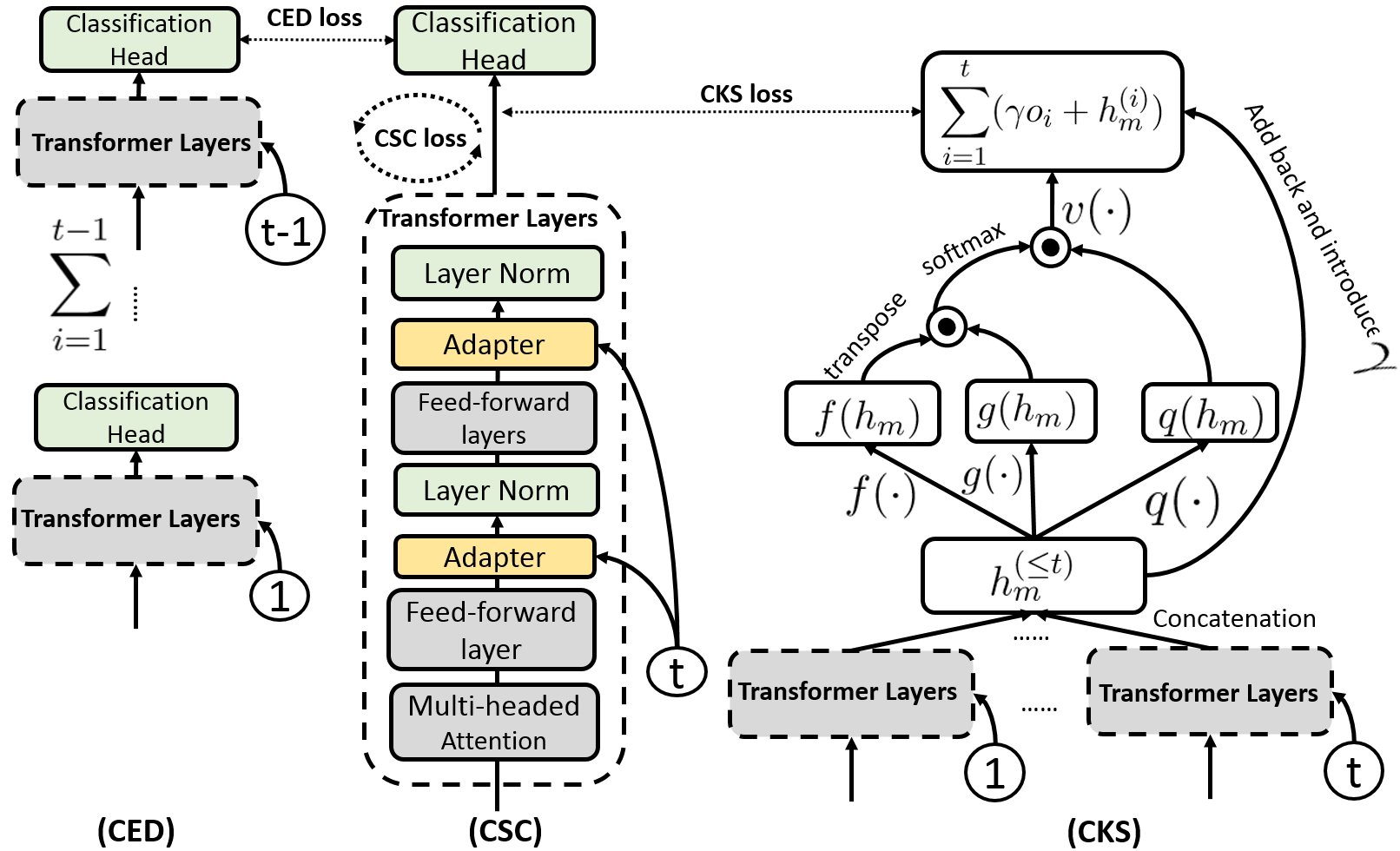}
\caption{
CLASSIC adopts Adapter-BERT \cite{DBLP:conf/icml/HoulsbyGJMLGAG19} and its adapters (yellow boxes) in a transformer~\cite{vaswani2017attention} layer ({\color{black}above (CSC))}. An adapter is a 2-layer fully connected network with a skip-connection. It is added twice to each Transformer layer. Only the adapters and layer norm (green boxes) layers are trainable. The other modules (grey boxes) {\color{black}of BERT are frozen. \textbf{(CSC):}} CSC loss is computed based on the current task model (details in Sec.~\ref{sec.csc_loss}). \textbf{(CED):} CED loss is computed based on all previous tasks from $1$ to $t-1$ (details in Sec. \ref{sec.ced_loss}). \textbf{(CKS):} CKS loss is computed based on previous and current tasks and a task-based self-attention. Details are given in Sec. \ref{sec.cks_loss}.
}
\label{overview_contrastive}
\vspace{-1mm}
\end{figure}

\textbf{BERT Adapter.} The idea was given in Adapter-BERT~\cite{DBLP:conf/icml/HoulsbyGJMLGAG19}, which inserts two 2-layer fully-connected networks (adapters) in each transformer layer of BERT 
(Figure~\ref{overview_contrastive}(CSC)). During training for the end-task, only the adapters and normalization layers are updated. All the other BERT parameters are frozen. This is good for CL as fine-tuning the BERT causes serious forgetting. Adapter-BERT achieves similar accuracy to the fine-tuned BERT \cite{DBLP:conf/icml/HoulsbyGJMLGAG19}. 

\subsection{Overview of CLASSIC}



{\color{black}
The architecture of CLASSIC is given in Figure~\ref{overview_contrastive}, which works in the DIL setting for ASC.} {It uses Adapter-BERT to avoid fine-tuning BERT. CLASSIC takes two inputs in training: (1) hidden states $h^{(t)}$ from the feed-forward layer of a transformer layer of BERT and 
(2) task id $t$ (no task id is needed in testing, see Sec.~\ref{sec.ced}). The outputs are hidden states with features for task $t$ to build a classifier. 

{\color{black}CLASSIC uses three sub-systems to achieve its objectives (see Sec.~\ref{sec.intro}):} 
(1) contrastive ensemble distillation (\textbf{CED}) for mitigating CF by distilling the knowledge of previous tasks to the current task model; (2) contrastive knowledge sharing (\textbf{CKS}) to encourage knowledge transfer; and (3) contrastive supervised learning on the current task model (\textbf{CSC}) to improve 
the current task model accuracy. 
We call this framework \textbf{contrastive continual learning}, inspired by contrastive learning. 

Contrastive learning uses multiple views of the existing data for representation learning to group similar data together and push dissimilar data far away, which makes it easier 
to learn a more accurate classifier. It uses various transformations of the existing data to create useful views. Given a mini-batch of $N$ training examples, if we create another view for each example, the batch will have $2N$ examples. We assume that $i$ and $j$ are two views of {\color{black}the training example.} 
If we use $i$ as \textbf{\textit{the anchor}}, $(i, j)$ is called a positive pair. All other pairs $(i, k)$ for $k \ne i$ are negative pairs. The contrastive loss for this positive pair is~\cite{chen2020simple}, 
\begin{equation}
\mathcal{L}_{i,j} = 
-\log\frac{\exp((h_i\cdot h_j)/\tau)}{\sum^{2N}_{k=1}\mathds{1}_{k\neq j}\exp((h_{i}\cdot h_{k})/\tau)},
\label{eq.contrastive}
\end{equation}
where the dot product $h_i\cdot h_j$ is regarded as a similarity function in the hidden space and $\tau$ is temperature. The final loss for the batch is calculated across all positive pairs. 
Eq.~\ref{eq.contrastive} is for unsupervised contrastive learning. It can also be used for supervised contrastive learning, where any two instances/views from the same class form a positive pair, and any instance of a class and any instance from other classes form a negative pair. 



}



\subsection{Overcoming Forgetting via Contrastive Ensemable Distillation (CED)}
\label{sec.ced_loss}

{\color{black}The CED objective is to deal with CF. 
We first introduce task masks that CED relies on to preserve the previous task knowledge/models to be distilled to the new task model to avoid CF.}

\begin{figure}[t]
\centering
\includegraphics[width=\columnwidth]{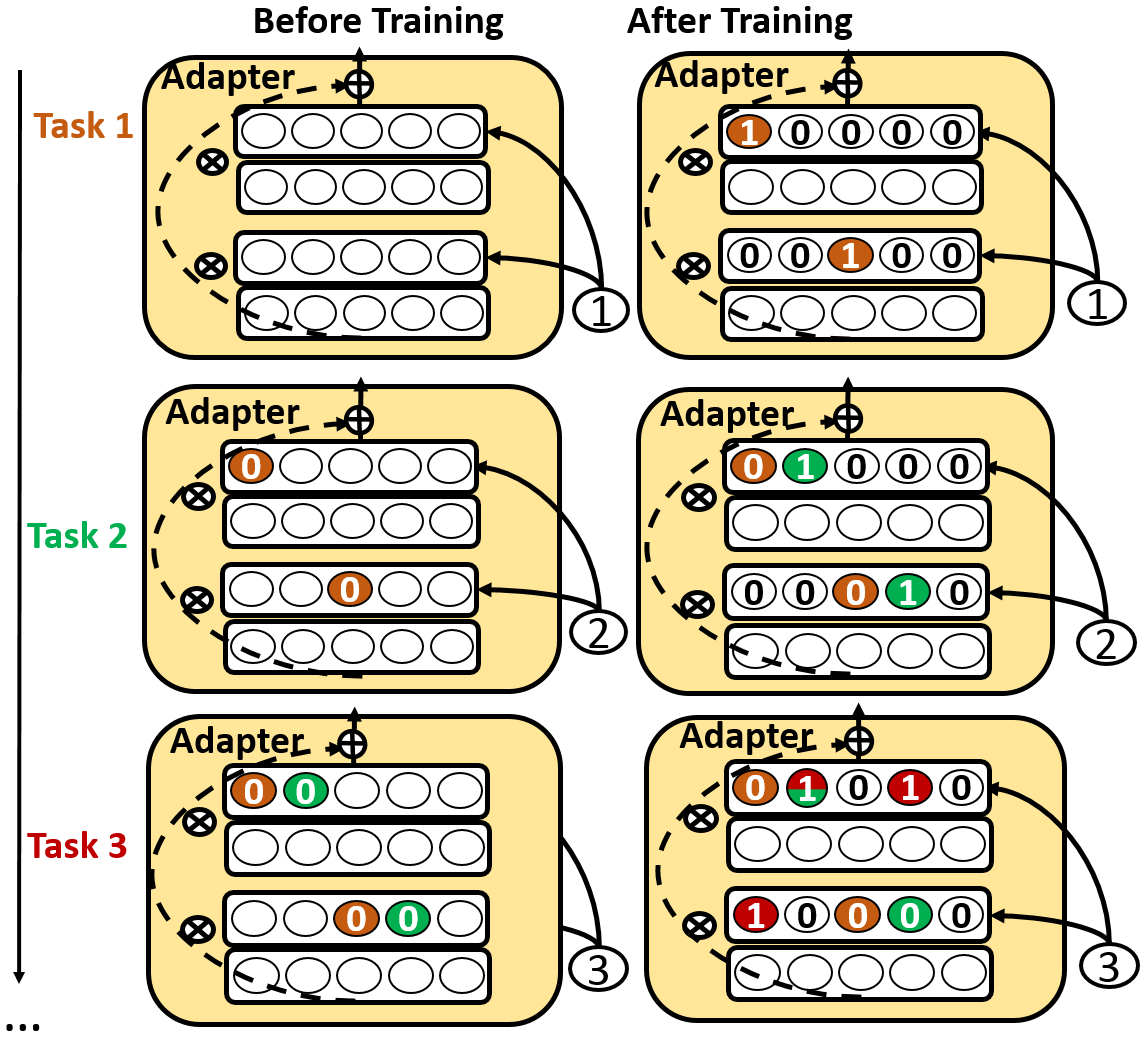}
\caption{Illustration of task masking: a (learnable) task mask is applied after the activation function to \textit{selectively} activate a neuron (or feature). The four rows of each task corresponds to the two fully-connected layers and their corresponding task masks. In the neurons before training, those with 0’s are the neurons to be protected (masked) and those
neurons without a number are free neurons (not used). In the neurons after training, those with 1’s show neurons that are important for the current task, which are used as masks for the future. Those neurons with more than one color indicate that they are shared by more than one task. Those 0 neurons without a color are not used by any task.}
\label{adapter_task_mask}
\vspace{-1mm}
\end{figure}

\subsubsection{Task Masks (TMs)}
Given the input hidden states $h^{(t)}$ from the feed-forward layer of a transformer layer, the adapter maps them into input $k_l^{(t)}$ via a fully-connected network, where $l$ is the $l$-th layer of the adapter. 
A TM (a ``soft'' binary mask) $\text{m}^{(t)}_l$ is trained for each task $t$ at each layer $l$ in the adapter during training task $t$'s classifier, indicating the neurons that are important for the task in the layer. Here we borrow the hard attention idea in \cite{Serra2018overcoming} and leverage the task id embedding to train the TMs.

For a task id $t$, its embedding $e^{(t)}_l$ consists of differentiable 
parameters that can be learned together with other parts of the network and it is trained for each layer in the adapter. 
To generate the TM $\text{m}^{(t)}_l$ from $e^{(t)}_l$, \textit{Sigmoid} is used as a pseudo-gate and a positive 
scaling hyper-parameter $s$ is applied to help training. The $m^{(t)}_l$ is computed as follows:
\vspace{-1.5mm}
\begin{equation}
\vspace{-1.5mm}
\label{eq1}
m^{(t)}_l = \sigma(se^{(t)}_l).
\end{equation}
Note that the neurons in $m^{(t)}_l$ may overlap with those in other $m^{(i_{\text{prev}})}_l$s from previous tasks showing some shared knowledge. Given the output of each layer in the adapter, $k_l^{(t)}$, we element-wise multiply $k_l^{(t)} \otimes m^{(t)}_l$. The masked output of the last layer $k^{(t)}$ is fed to the next layer of the BERT with a skip-connection (see Figure~\ref{adapter_task_mask}) 
After learning task $t$, the final $m^{(t)}_l$ is saved and added to the set $\{m^{(t)}_l\}$. 

\subsubsection{Training Task Masks (TMs)} 
For each previous task $i_{\text{prev}} \in \mathcal{T}_{\text{prev}}$, its TM $m^{(i_{\text{prev}})}_l$ 
indicates which neurons are used by that task and need to be protected. 
In learning task $t$, $m^{(i_{\text{prev}})}_l$ is used to set the gradient $g^{(t)}_l$ on \textit{all} used neurons of the layer $l$ to 0. 
Before modifying the gradient, we first accumulate all used neurons by all previous tasks TMs.
Since $m^{(i_{\text{prev}})}_l$ is binary, we use 
max-pooling 
to achieve the accumulation:  
\vspace{-1.5mm}
\begin{equation}
\vspace{-1.5mm}
m^{(t_{\text{ac}})}_l = \text{MaxPool}(\{m^{(i_{\text{prev}})}_l\}).
\end{equation}
The term $m^{(t_{\text{ac}})}_l$ is applied to the gradient:
\vspace{-1.5mm}
\begin{equation}
\vspace{-1.5mm}
g^{'(t)}_l = g^{(t)}_l \otimes (1-m^{(t_{\text{ac}})}_l).
\end{equation}
Those gradients corresponding to the 1 entries in $m^{(t_{\text{ac}})}_l$ are set to 0 while the others remain unchanged. 
In this way, neurons in an old task are protected. 
Note that we expand (copy) the vector $m_l^{(t_{\text{ac}})}$ to match the dimensions of $g_l^{(t)}$.

Though the idea is intuitive, $e^{(t)}_l$ is not easy to train. To make the learning of $e^{(t)}_l$ easier and more stable, an annealing strategy is applied~\cite{Serra2018overcoming}. That is, $s$ is annealed during training, inducing a gradient flow and set {\color{black}$s=s_{\max}$} during testing. {\color{black}Eq.~\ref{eq1} approximates a unit step function as the mask, with $m_l^{(t)} \to \{0, 1\}$ when $s \to \infty$. 
A training epoch starts with all neurons being equally active, 
which are progressively polarized within the epoch. Specifically, $s$ is annealed as follows}:
\vspace{-1.5mm}
\begin{equation}
\vspace{-1.5mm}
\label{eq:smax}
s = \frac{1}{s_{\max}} + (s_{\max} - \frac{1}{s_{\max}})\frac{b-1}{B-1},
\end{equation}
where $b$ is the batch index and $B$ is the total number of batches in an epoch.

\textbf{Illustration.} In Figure~\ref{adapter_task_mask}, after learning Task 1, we obtain its useful neurons marked in orange with a ``1'' in each neuron, which serves as a mask in learning future tasks. In learning Task 2, 
those useful neurons for Task 1 are masked (with ``0'' in those orange neurons on the left). The process also learns the useful neurons for Task 2 marked in green with ``1''s. When Task 3 arrives, all neurons for Tasks 1 and 2 are masked, i.e., its TM entries are set to 0 (orange and green before training). After training Task 3, we see that Task 3 and Task 2 have a shared neuron that is important to both. The shared neuron is marked in both red and green.

\subsubsection{Contrastive Ensemble Distillation (CED)}
\label{sec.ced}

The TMs mechanism isolates different parameters for different tasks. This seems to be perfect for overcoming forgetting since the previous task parameters are fixed and cannot be updated by future tasks. However, since DIL setting does not have task id in testing, we cannot directly take the advantage of the TMs. To address this issue, we propose the CED objective to help distill \textit{all} previous knowledge to the current task model so that we can simply use \textit{the last model} as the final model without requiring the task id in testing. 

\textbf{Representation of Previous Tasks.} Recall that we know which neurons/units are for which task $i$ by reading $\{m^{(i)}_l\}$. For each previous task $i$ of the current task $t$, we can compute its masked output of Adapter-BERT $h^{(i)}_{m}$ (the layer before the classification head) by applying $m^{(i)}_l$ to the Adapter-BERT.

\textbf{Ensemble Distillation Loss.}~We distill the knowledge of the ensemble of previous tasks into the single current task model. As we have a shared classification head for all tasks in DIL, which is exposed to forgetting, the distillation should be based on the output of the classification head. Specifically, given a previous task's Adapter-BERT output $h^{(i)}_{m}$, we compute the output of the classification head using $h^{(i)}_{m}$, which gives us the logit {(un-normalized prediction)} value $z^{(i)}_{m}$. {\color{black}We then distill the knowledge using $z^{(i)}_{m}$ and the current task classification head output $z^{(t)}_{m}$ based on contrastive loss}, inspired by 
\cite{DBLP:conf/iclr/TianKI20},
\begin{equation}
\small
\mathcal{L}_{\text{CED}}^{(i)} = \sum_{n=1}^{2N}-\log\frac{\exp((z^{(i)}_{m:2n-1}\cdot z^{(t)}_{m:2n})/\tau)}{\sum^{2N}_{j=1}\mathds{1}_{n\neq j}\exp((z^{(i)}_{m:n}\cdot z^{(t)}_{m:j})/\tau)},
\end{equation}
where $N$ is the batch size and $\tau>0$ is an adjustable temperature parameter controlling the separation of classes. {\color{black}The index $n$ is the \textit{anchor} and the notation $z^{(i)}_{m:n}$ refers to the $n$-th sample in $z^{(i)}_{m}$. $z^{(i)}_{m:2n-1}$ and $z^{(t)}_{m:2n}$ are the logits of previous and current task models for the same input sample, a \textit{positive} pair in contrastive learning. All the other possible pairs are \textit{negative pairs}. Note that for each anchor $i$, there is $1$ positive pair and $2N-2$ negative pairs. The denominator has a total of $2N-1$ terms (both the positives and negatives)}.  Note that the previous task models are fixed and thus can serve as teacher networks. As we have $i\ge1$ previous tasks, hence $i\ge1$ teacher networks but only one current task student network. We adopt the contrastive framework by defining multiple pair-wise contrastive losses between $z^{(i)}_{m}$ and $z^{(t)}_{m}$. These losses are summed up to give the final CED loss, 
\begin{equation}
\mathcal{L}_{\text{CED}} = \sum_{i=1}^{t-1}\mathcal{L}_{\text{CED}}^{(i)}.
\end{equation}

\subsection{Transferring Knowledge via Contrastive Knowledge Sharing (CKS)}
\label{sec.cks_loss}

{\color{black}
CKS aims to capture the shared knowledge among tasks and help the new task learn a better representation and better classifier. 
The intuition of CKS is as follows: {Contrasive learning has the ability to capture the shared knowledge between different views \cite{DBLP:conf/nips/Tian0PKSI20,DBLP:journals/corr/abs-1807-03748}. This is achieved by seeking representation that are invariant cross similar views. 
} 
If we can generate a view from previous tasks that is similar to the current task, 
the contrastive loss can capture the shared knowledge and learn a representation for knowledge transfer to the new task learning. 
Below, we first introduce how to construct such a view and use it in the CKS objective.}

\subsubsection{Task-based Self-Attention}

Intuitively, the more similar the two tasks are, the more shared knowledge they have. To achieve our goal, we should combine all similar tasks as the shared knowledge view. In order to focus on the similar tasks, we propose to use task-based self-attention mechanism to attend to them. Inspired by \cite{DBLP:journals/corr/abs-1805-08318}, given the concatenation of the output of Adapter-BERT for all previous and current tasks, {\color{black} $h^{(\leq t)}_{m} = \text{cat(} \{h^{(i)}_{m}\}_{i=1}^{t}\text{)}$, and task $i\leq t$, we first transform it into two feature spaces via 
$f(h^{(i)}_{m})=W_fh^{(i)}_{m}$, $g(h^{(i)}_{m})=W_gh^{(i)}_{m}$ (see Figure~\ref{overview_contrastive}(CKS)).} 

To compare the similarity between tasks $i\leq t$ and $j\leq t$, 
we calculate similarity $s_{ij}$ via
\begin{equation}
s_{ij} = f(h^{(i)}_{m})^Tg(h^{(j)}_{m}).
\end{equation}
We then compute the attention score $\alpha_{j,i}$ to indicate which similar tasks ({similar to the current task $t$}) should be attended to based on the current task data,
\begin{equation}
\alpha_{j,i} = \frac{\exp(s_{ij})}{\sum_{i=1}^t\exp(s_{ij})}.   
\end{equation} 
The attention score is applied to each task in $h^{(\leq t)}_{m}$ to get the attention output $o_j$ using weighted sum:
\begin{equation}
o_{j} = v(\sum^t_{i=1}\alpha_{j,i}q(h^{(i)}_{m})),
\end{equation}
where $v(\cdot)$ and $q(\cdot)$ are two functions for transforming feature spaces: $v(h^{(i)}_{m})=W_vh^{(i)}_{m}$ and $q(h^{(i)}_{m})=W_qh^{(i)}_{m}$. 

Lastly, we multiply the output of the attention layer by a scale parameter and add back to the input feature $h^{(\leq t)}_{m}$. The final output of the $h^{(\leq t)}_{\text{CKS}}$ is the sum over all considered tasks,
\begin{equation}
h^{(\leq t)}_{\text{CKS}} = \sum^t_{i=1}(\gamma o_{i} + h^{(i)}_{m}),
\end{equation}
where $\gamma$ is a learnable scalar and it is initialized to 0. This allows the model to first learn on the current task and then gradually learn to assign more weights to other tasks.

\subsubsection{Knowledge Sharing Loss}

The output of the task-based self-attention provides us the knowledge sharing view $h^{(\leq t)}_{\text{CKS}}$. Along with the output of Adapter-BERT for the current task $h^{(t)}_{m}$, we can easily perform contrastive learning between these two views. Note that $h^{(\leq t)}_{\text{CKS}}$ is computed based on the current task data and their corresponding class labels, so we give the two views have the same label and thus we can integrate the label information in our CKS loss, 
\begin{equation}
\small
\begin{aligned}
\mathcal{L}_{\text{CKS}} = &\sum_{n=1}^{N}-\frac{1}{N_{y_n}-1}\sum_{j=1}^N\mathds{1}_{n\leq j}\mathds{1}_{y_n=y_j}\\
&\log\frac{\exp((h^{(\leq t)}_{\text{CKS}:n}\cdot h^{(t)}_{m:j})/\tau)}{\sum^{N}_{k=1}\mathds{1}_{n\neq k}\exp((h^{(\leq t)}_{\text{CKS}:n}\cdot h^{(t)}_{m:k})/\tau)},
\end{aligned}
\end{equation}
where $N$ is the batch size and $N_{y_n}$ is the number of examples in the batch that have the label $y_n$. {$h^{(\leq t)}_{\text{CKS}}$ is the first view while $h^{(t)}_{m}$ is the second view. The shared knowledge between them represents the shared knowledge between previous and current tasks. Different from the CED loss, the CKS loss leverages the class information and thus can have multiple positive pairs decided by whether two samples share the same class label.}

\subsection{Contrastive Supervised Learning of the Current Task (CSC)}
\label{sec.csc_loss}

We further improve the performance of the current task by adopting the supervised contrastive loss \cite{DBLP:journals/corr/abs-2004-11362} on the \textit{current task} $h^{(t)}_{m}$, 
\begin{equation}
\small
\begin{aligned}
\mathcal{L}_{\text{CSC}} = &\sum_{n=1}^{N}-\frac{1}{N_{y_n}-1}\sum_{j=1}^N\mathds{1}_{n\leq j}\mathds{1}_{y_n=y_j}\\
&\log\frac{\exp((h^{(t)}_{m:n}\cdot h^{(t)}_{m:j})/\tau)}{\sum^{N}_{k=1}\mathds{1}_{n\neq k}\exp((h^{(t)}_{m:n}\cdot h^{(t)}_{m:k})/\tau)}.
\end{aligned}
\end{equation}

\subsection{Final Loss}
The final loss is the weighted average of the supervised cross entropy (CE) loss, CSC loss, and the proposed CED and CKS losses:
\begin{equation}
\label{eq.13}
\mathcal{L} = \mathcal{L}_{\text{CE}} + \lambda_{\text{1}}\mathcal{L}_{\text{CSC}} + \lambda_{\text{2}}\mathcal{L}_{\text{CED}} + \lambda_{\text{3}}\mathcal{L}_{\text{CKS}}.  
\end{equation}

\section{Experiments}

{\color{black} This section evaluates the proposed CLASSIC system and compares it with both \textit{non-continual learning} and \textit{continual learning} baselines.}

\subsection{Experiment Datasets}

\begin{table}[]
\centering
\resizebox{1.0\columnwidth}{!}{
\begin{tabular}{ccccc}
\specialrule{.2em}{.1em}{.1em}
Data source & Task/domain & Train & Validation & Test \\
\specialrule{.1em}{.05em}{.05em}

\multirow{3}{*}{Liu3domain} & Speaker & 352  & 44 & 44 \\
 & Router & 245 & 31 & 31 \\
 & Computer & 283 & 35 & 36  \\
\specialrule{.1em}{.05em}{.05em}
\multirow{5}{*}{HL5domain} & Nokia6610 & 271 & 34 & 34  \\
 & Nikon4300 & 162 & 20  & 21 \\
 & Creative & 677 & 85 & 85 \\
 & CanonG3 & 228 & 29 & 29 \\
 & ApexAD & 343 & 43 & 43 \\
\specialrule{.1em}{.05em}{.05em}
\multirow{9}{*}{Ding9domain} & CanonD500 & 118 & 15 & 15 \\
 & Canon100 & 175 & 22 & 22 \\
 & Diaper & 191 & 24 & 24 \\
 & Hitachi & 212 & 26 & 27\\
 & Ipod & 153 & 19 & 20 \\
 & Linksys & 176 & 22 & 23 \\
 & MicroMP3 & 484 & 61 & 61 \\
 & Nokia6600 & 362 & 45 & 46 \\
 & Norton & 194 & 24 & 25 \\
\specialrule{.1em}{.05em}{.05em}
\multirow{2}{*}{SemEval14} & Rest. & 3452 & 150 & 1120 \\
 & Laptop & 2163 & 150 & 638 \\
\specialrule{.1em}{.05em}{.05em}

\end{tabular}
}
\caption{Number of sentences in each task or dataset. More detailed statistics are given in \textit{Supplementary}.}
\label{tab:dataset_main}
\vspace{-1mm}
\end{table}
{\color{black}
We use 19 ASC datasets to produce sequences of 19 tasks. Each dataset is a set of {\color{black}aspect and sentiment annotated review sentences} from reviews of a particular product and represents a task. The datasets are from 4 sources: (1) \textbf{HL5Domains} \cite{hu2004mining}: review sentences of 5 products; (2) \textbf{Liu3Domains} \cite{liu2015automated}: review sentences of 3 products; (3) \textbf{Ding9Domains} \cite{ding2008holistic}: review sentences of 9 products; and (4) \textbf{SemEval14}: review sentences of 2 products - SemEval 2014 Task 4 for laptop and restaurant. 
{\color{black}To be consistent with the existing research \cite{tang-etal-2016-aspect},
sentences with both positive and negative sentiments about an aspect are not used.}
Statistics of the 19 datasets are given in Table \ref{tab:dataset_main}.

\subsection{Compared Baselines}
\label{sec:baselines}



We employ \textbf{46} baselines, which include both \textit{non-continual learning} and \textit{continual learning} methods. {\color{black}Since little work has been done in DIL, we adapt the recent TIL systems to DIL by merging classification heads to form DIL systems.} 

\textbf{Non-Continual Learning Baselines}: Each of these baselines builds a separate model for each task independently, which we call a ONE variant. It thus has no knowledge transfer or CF. There are 8 ONE variants. Four are created using \textbf{(1) BERT} with fine-tuning, (2) \textbf{BERT (Frozen)} without fine-tuning \textbf{(3) Adapter-BERT}~\cite{DBLP:conf/icml/HoulsbyGJMLGAG19} and \textbf{(4) W2V} (word2vec embeddings trained with the Amazon review data in \cite{Xu2018pro} using FastText \cite{grave2018learning}). Adding CSC (Contrastive Supervised learning of the Current task) creates another 4 variants. We adopt the ASC network in~\cite{DBLP:conf/acl/LiX18}, taking aspect term and review sentence as input for BERT variants. For W2V variants, we use their concatenation. 

\textbf{Continual Learning (CL) Baselines}. The CL setting has 38 baselines in 5 categories. The first category uses a naive CL (NCL) approach. It simply uses a network to learn all tasks with no mechanism to deal with CF or knowledge transfer. Like ONE, we have 8 NCL variants. The second category has 11 baselines created using recent CL methods \textbf{KAN}~\cite{ke2020continual}, \textbf{SRK}~\cite{DBLP:conf/dasfaa/LvWLCZ19}, \textbf{HAT}~\cite{Serra2018overcoming}, \textbf{UCL}~\cite{DBLP:conf/nips/AhnCLM19}, \textbf{EWC}~\cite{Kirkpatrick2017overcoming}, \textbf{OWM}~\cite{zeng2019continuous} and \textbf{DER++}~\cite{buzzega2020dark}.  
KAN and SRK are for document sentiment classification. We use the concatenation of the aspect and the sentence as input. 
HAT, UCL, EWC, OWM and DER++ were originally designed for image classification. We replace their original image classification networks with CNN for text classification \cite{DBLP:conf/emnlp/Kim14}. 
{\color{black}{HAT} is one of the best TIL methods with almost no forgetting. 
{UCL} is a recent TIL method. 
{EWC} is a popular CIL method, which was adapted 
for TIL in~\cite{Serra2018overcoming}. They are converted to DIL versions by merging their classification heads.
{OWM} \cite{zeng2019continuous} is a CIL method, which we also adapt to a DIL method like EWC. DER++ and SRK can work in the DIL setting. HAT and KAN require task id as an input in testing and cannot function in the DIL setting.} We create two variants of HAT (and KAN): using the \textit{last} model in testing as CLASSIC does or detecting task id using the entropy method \textit{ent} in~\cite{von2019continual}. This category uses BERT (Frozen) as the base. The third category has 7 baselines using Adapter-BERT. KAN and SRK cannot be adapted to use adapters. The fourth category uses W2V, which gives another 11 baselines. The final category has one baseline \textbf{LAMOL} \cite{DBLP:conf/iclr/SunHL20}, which uses the GPT-2 model.

}


\vspace{+1mm}
\noindent
{\color{black}\textbf{Evaluation Protocol:} We follow the standard CL evaluation method in~\cite{DBLP:journals/corr/abs-1909-08383}.
We first present CLASSIC a sequence of ASC tasks for it to learn. Once a task is learned, its training data is discarded. After all tasks are learned, we test using the test data of all tasks without giving task ids. }

\subsection{Hyperparameters}
{\color{black}Unless otherwise stated, the adapter uses 2 layers of fully connected network with dimensions 2000. The task id embeddings have 2000 dimensions. A fully connected layer with softmax output is used as the classification head in the last layer of BERT. We use 400 for $s_{\max}$ in Eq.~\ref{eq:smax},
dropout of 0.5 between fully connected layers. The temperature $\tau$ in each contrastive objective is set to 1 {\color{black}(see Supplementary for parameter tuning)}. The weight of each objective in Eq.~\ref{eq.13} is set to 1. We use the embedding of \texttt{[CLS]} as the output of Adapter-BERT. For CKS and CSC, we use $l_2$ normalization on the output of Adapter-BERT before computing the contrastive loss. The training of BERT, Adapter-BERT and CLASSIC follow that of \cite{DBLP:conf/naacl/XuLSY19}. We adopt $\text{BERT}_{\textbf{BASE}}$ (uncased). The max length of the sum of sentence and aspect is 128. We use Adam optimizer and set the learning rate to 3e-5.
For the SemEval datasets, 10 epochs are used and for all other datasets, 30 epochs are used based on results from validation data.
All runs use the batch size 32. For CL baselines, we train all models with the learning rate of 0.05, early-stop training when there is no improvement in the validation loss for 5 epochs and set the batch size to 64. We use the code provided by their authors and adopt their original parameters (for EWC, we adopt the variant
implemented by \cite{Serra2018overcoming}).
}



\begin{table}[h!]
\centering
\resizebox{\columnwidth}{!}{
\begin{tabular}{ccc||cc}
\specialrule{.2em}{.1em}{.1em}
Scenario & Category & Model  & Acc. & MF1 \\
\specialrule{.1em}{.05em}{.05em}
\multirow{8}{*}{\begin{tabular}[c]{@{}c@{}}Non-continual\\      Learning\end{tabular}} 
& \multirow{2}{*}{BERT} & ONE & 0.8584 & 0.7635 \\
 &  & ONE+csc  &  0.8353	& 0.7388  \\
  & \multirow{2}{*}{BERT (Frozen)} & ONE & 0.7814 & 0.5813 \\
 &  & ONE+csc &  0.8265 & 0.7232  \\
 & \multirow{2}{*}{Adapter-BERT} & ONE & 0.8596 & 0.7807 \\
  &  & ONE+csc &  0.8530 & 0.7516  \\
 & \multirow{2}{*}{W2V} & ONE & 0.7701 & 0.5189 \\
 &  & ONE+csc &  0.7761	& 0.5487 \\
 \cline{1-5}
\multirow{40}{*}{\begin{tabular}[c]{@{}c@{}}Continual\\      Learning\end{tabular}} 
& \multirow{2}{*}{BERT} & NCL & 0.8048 & 0.7085  \\
 &  & NCL+csc  &  0.7727 & 0.5807 \\
& \multirow{2}{*}{BERT (Frozen)} & NCL & 0.8685 & 0.7873 \\
 &  & NCL+csc  & 0.8693 & 0.7912 \\
 & \multirow{2}{*}{Adapter-BERT} & NCL & 0.8667 & 0.7804 \\
  &  & NCL+csc  & 0.8809 & 0.7847  \\
 & \multirow{2}{*}{W2V} & NCL & 0.8408	
 & 0.7455 \\
  &  & NCL+csc &  0.8396 & 0.7509\\

 \cline{2-5}
 & \multirow{10}{*}{\begin{tabular}[c]{@{}c@{}}BERT\\      (Frozen)\end{tabular}} & KAN & \multicolumn{2}{c}{---}\\
 &  & KAN+last & 0.8320 & 0.7352 \\
 &  & KAN+ent &  0.8278 & 0.7243 \\
 &  & SRK  & 0.8391 & 0.7438 \\
 &  & EWC  & 0.8660 & 0.7831 \\
 &  & UCL & 0.8538 & 0.7690 \\
 &  & OWM & 0.8611 & 0.7665 \\
 &  & DER++ & 0.8753 & 0.8009 \\
 &  & HAT  & \multicolumn{2}{c}{---}\\
 &  & HAT+last &  0.8473 & 0.7649 \\  
 &  & HAT+ent &  0.8418 & 0.7614  \\
 \cline{2-5}
 
  & \multirow{6}{*}{Adapter-BERT} &  EWC & 0.8805 & 0.7875 \\
 &  & UCL & 0.7123 & 0.3961 \\
 &  & OWM & 0.8766 & 0.7882 \\
 &  & DER++ & 0.8859 & 0.7985 \\
 &  & HAT & \multicolumn{2}{c}{---} \\
 &  & HAT+last & 0.8823 & 0.7919 \\  
 &  & HAT+ent & 0.8854 & 0.8245 \\
 \cline{2-5}
 
 & \multirow{10}{*}{W2V} & KAN & \multicolumn{2}{c}{---} \\	
 &  & KAN+last *  & 0.7123 & 0.3961 \\  
 &  & KAN+ent *  &  0.7123 & 0.3961  \\
 &  & SRK * & 0.7123 & 0.3961 \\
 &  & EWC &  0.7586	& 0.6545  \\
 &  & UCL & 0.8187 & 0.6965 \\
 &  & OWM & 0.8256 & 0.7253 \\
 &  & DER++ & 0.8459 & 0.7722 \\
 &  & HAT & \multicolumn{2}{c}{---}\\
 &  & HAT+last &  0.7599 & 0.5849 \\  
 &  & HAT+ent  &  0.7605 & 0.5349 \\
 \cline{2-5}
 & \multicolumn{2}{c||}{LAMOL} & 0.8891 & 0.8059 \\
 \cline{2-5}
 & \multicolumn{2}{c||}{{CLASSIC (forward)}}  & 0.8886 & 0.8365 \\
 & \multicolumn{2}{c||}{\textbf{CLASSIC}} & \textbf{0.9022} & \textbf{0.8512} \\
\specialrule{.1em}{.05em}{.05em}
\vspace{-1mm}
\end{tabular}
}
\caption{Accuracy~(Acc.)~and~Macro-F1~(MF1)~averaged over 5 random sequences of 19 tasks.~KAN and HAT need task id in testing and thus have no results.~KAN and SRK (RNN based) cannot work with Adapters.~{\color{black}\textbf{*}:~KAN and SRK under W2V fail to train.} \textbf{Standard deviation} showing {statistical significance}, \textbf{network size} and \textbf{running time} are in Supplementary. 
\vspace{-5mm}
} 
\label{tab:overall_results}
\end{table}

\subsection{Results and Analysis}
\label{sec:results}


As the order of the 19 tasks can influence the final results, we randomly select and run 5 task sequences and report their average results in Table~\ref{tab:overall_results}.
We compute both accuracy and Macro-F1, 
where Macro-F1 is the main metric as the imbalanced classes introduce biases in accuracy. 

\textbf{Overall}, Table~\ref{tab:overall_results} shows that CLASSIC outperforms all baselines markedly. 

\textbf{(1)}.~For non-continual learning baselines (ONE variants), Adapter-BERT performs similarly to BERT (fine-tuning). Both BERT (Frozen) and W2V variants are weaker, which is understandable.  

\textbf{(2).} Comparing ONE variants and NCL variants, we see that under W2V, NCL variants are much better than ONE variants. This indicates ASC tasks are similar and have shared knowledge. Catastrophic forgetting (CF) is not a major issue for W2V.

However, BERT NCL (fine-tuning) is much worse than BERT ONE and Adapter-BERT NCL (adapter-tuning) as BERT fine-tuning learns highly task specific knowledge \cite{DBLP:journals/corr/abs-2004-14448}. While this is desirable for ONE, it is bad for NCL because task specific knowledge is hard to share across tasks, which causes forgetting (CF). The +csc options are poor for BERT ONE and NCL.

\textbf{(3).} Various continual learning (CL) baselines with BERT (Frozen) are also markedly weaker than CLASSIC. Baselines that can use Adapter-BERT are also much poorer than CLASSIC. Note that SRK and KAN cannot work with Adapter-BERT.

\textbf{(4).} W2V based CL baselines are even weaker.

\textbf{(5).} Since both KAN and HAT need task id in testing and the DIL setting does not provide task id, they have no results. 
But we use the last model (+last) or use an existing entropy-based method (+ent)~\cite{von2019continual} to automatically identify the task id for each test instance. These variants are also markedly weaker than CLASSIC. 

\textbf{(6).} LAMOL is based on GPT-2 and its performance is weaker than CLASSIC too.





\textbf{Effectiveness of Knowledge Transfer.}
The results under CLASSIC(forward) in Table~\ref{tab:overall_results} are the average results computed using the accuracy/MF1 of each task when it was first learned. The results under CLASSIC are the final average results after all tasks are learned, including backward transfer. By comparing ONE variants and CLASSIC(forward), we can see whether forward transfer is effective. By comparing CLASSIC(forward) and CLASSIC, we can see whether the backward transfer can improve further. We see both forward and backward transfers are effective.

\subsection{Ablation Experiments}

The results of ablation experiments are given in Table \ref{tab:ablation_results}. ``-CKS'', ``-CSC'' and ``-CED'' mean without constrastive knowledge sharing,  contrastive supervised learning on the current task and contrastive ensemble distillation, respectively. Table \ref{tab:ablation_results} clearly shows that each of the components is effective and they work in concert to produce the best final result. 


\begin{table}[]
\centering
\resizebox{0.7\columnwidth}{!}{
\begin{tabular}{l||cc}
\specialrule{.2em}{.1em}{.1em}
Model & Acc. & MF1 \\
\specialrule{.1em}{.05em}{.05em}
\textbf{CLASSIC}  & \textbf{0.9022} & \textbf{0.8512} \\
\hline
-CSC & 0.8872 & 0.8007 \\
-CKS & 0.8915 & 0.8232   \\
-CED & 0.8828 & 0.7934 \\
-CKS,-CED &  0.8864 & 0.7969  \\
-CKS,-CSC & 0.8926 & 0.8346 \\
-CED,-CSC & 0.8868 & 0.8032 \\
-CED,-CKS,-CSC  & 0.8823 & 0.7919 \\
\specialrule{.1em}{.05em}{.05em}
\end{tabular}
}
\caption{Ablation experiment results. 
}

\label{tab:ablation_results}
\vspace{-1mm}
\end{table}


\section{Conclusion}

This paper studied \textit{domain incremental learning} (DIL) of a sequence of ASC tasks 
without knowing the task ids in testing. Our method CLASSIC uses \textit{adapters} to exploit BERT and to deal with BERT CF in fine-tuning, and the proposed \textit{contrastive continual learning} to transfer knowledge across tasks and to distill knowledge from previous tasks to the current task so that the last model can be used for all tasks in testing and no task id is needed. 
{\color{black}Our experimental results show that CLASSIC outperforms the state-of-the-art baselines.

Finally, we believe that the idea of CLASSIC is also applicable to some other NLP tasks. For example, in named entity extraction, we can build a better model to extract the same types of entities from text of different domains. Each domain works on the same task but no data sharing (the data may be from different clients with privacy concerns). Since this is an extraction task, the backbone model needs to be switched to an extraction model. } 


\section*{Acknowledgments}
This work was supported in part by two grants from National Science Foundation: IIS-1910424 and IIS-1838770, a DARPA Contract HR001120C0023, and a research gift from Northrop Grumman.





\bibliography{anthology,custom}
\bibliographystyle{acl_natbib}

\appendix

\section{Appendix}
\label{sec:appendix}

\subsection{Detailed Datasets Statistics}
\label{sec:dataset_detail}

Table 1 in the main paper has already showed the number of examples in each dataset. Here we provide additional details about aspects and sentiments. The detailed statistics of the 19 datasets or tasks are given in Table~\ref{tab:dataset} here.

\begin{table*}[]
\centering
\resizebox{\textwidth}{!}{
\begin{tabular}{ccccc}
\specialrule{.2em}{.1em}{.1em}
Dataset & Domains & Training & Validating & Testing \\
\specialrule{.1em}{.05em}{.05em}

\multirow{3}{*}{Liu3domain} & Speaker & 233 S./352 A./287 P./65 N./0 Ne. & 30 S./44 A./35 P./9 N./0 Ne. & 38 S./44 A./40 P./4 N./0 Ne. \\
 & Router & 200 S./245 A./142 P./103 N./0 Ne. & 24 S./31 A./19 P./12 N./0 Ne. & 22 S./31 A./24 P./7 N./0 Ne. \\
 & Computer & 187 S./283 A./218 P./65 N./0 Ne. & 25 S./35 A./23 P./12 N./0 Ne. & 29 S./36 A./29 P./7 N./0 Ne. \\
\specialrule{.1em}{.05em}{.05em}
\multirow{5}{*}{HL5domain} & Nokia6610 & 209 S./271 A./198 P./73 N./0 Ne. & 29 S./34 A./30 P./4 N./0 Ne. & 28 S./34 A./25 P./9 N./0 Ne. \\
 & Nikon4300 & 131 S./162 A./135 P./27 N./0 Ne. & 15 S./20 A./18 P./2 N./0 Ne. & 15 S./21 A./19 P./2 N./0 Ne. \\
 & Creative & 582 S./677 A./422 P./255 N./0 Ne. & 68 S./85 A./42 P./43 N./0 Ne. & 70 S./85 A./52 P./33 N./0 Ne. \\
 & CanonG3 & 190 S./228 A./180 P./48 N./0 Ne. & 25 S./29 A./21 P./8 N./0 Ne. & 24 S./29 A./24 P./5 N./0 Ne. \\
 & ApexAD & 281 S./343 A./146 P./197 N./0 Ne. & 35 S./43 A./16 P./27 N./0 Ne. & 28 S./43 A./31 P./12 N./0 Ne. \\
\specialrule{.1em}{.05em}{.05em}
\multirow{9}{*}{Ding9domain} & CanonD500 & 103 S./118 A./96 P./22 N./0 Ne. & 11 S./15 A./14 P./1 N./0 Ne. & 13 S./15 A./11 P./4 N./0 Ne. \\
 & Canon100 & 137 S./175 A./123 P./52 N./0 Ne. & 19 S./22 A./20 P./2 N./0 Ne. & 16 S./22 A./21 P./1 N./0 Ne. \\
 & Diaper & 166 S./191 A./143 P./48 N./0 Ne. & 22 S./24 A./18 P./6 N./0 Ne. & 24 S./24 A./22 P./2 N./0 Ne. \\
 & Hitachi & 152 S./212 A./153 P./59 N./0 Ne. & 23 S./26 A./19 P./7 N./0 Ne. & 23 S./27 A./14 P./13 N./0 Ne. \\
 & Ipod & 124 S./153 A./101 P./52 N./0 Ne. & 18 S./19 A./14 P./5 N./0 Ne. & 19 S./20 A./15 P./5 N./0 Ne. \\
 & Linksys & 152 S./176 A./128 P./48 N./0 Ne. & 19 S./22 A./13 P./9 N./0 Ne. & 20 S./23 A./16 P./7 N./0 Ne. \\
 & MicroMP3 & 384 S./484 A./340 P./144 N./0 Ne. & 42 S./61 A./48 P./13 N./0 Ne. & 51 S./61 A./39 P./22 N./0 Ne. \\
 & Nokia6600 & 298 S./362 A./244 P./118 N./0 Ne. & 26 S./45 A./32 P./13 N./0 Ne. & 39 S./46 A./30 P./16 N./0 Ne. \\
 & Norton & 168 S./194 A./54 P./140 N./0 Ne. & 17 S./24 A./15 P./9 N./0 Ne. & 24 S./25 A./5 P./20 N./0 Ne. \\
\specialrule{.1em}{.05em}{.05em}
\multirow{2}{*}{SemEval14} & Rest & 1893 S./3452 A./2094 P./779 N./579 Ne. & 84 S./150 A./70 P./26 N./54 Ne. & 600 S./1120 A./728 P./196 N./196 Ne. \\
 & Laptop & 1360 S./2163 A./930 P./800 N./433 Ne. & 98 S./150 A./57 P./66 N./27 Ne. & 411 S./638 A./341 P./128 N./169 Ne. \\
\specialrule{.1em}{.05em}{.05em}

\end{tabular}
}
\caption{Statistics of the datasets. \textbf{S}.: number of sentences; \textbf{A}: number of aspects; \textbf{P., N., and Ne.}: number
of positive, negative and neutral aspect polarities respectively. Note that SemEval14 has 3 classes of polarities while the others have only 2 classes (positive and negative) because in these datasets, those sentences without sentiment (neutral) are not annotated with aspects. Thus, we cannot use them for aspect sentiment classification (ASC). 
}
\label{tab:dataset}
\end{table*}

\begin{table}[]
\centering
\resizebox{\columnwidth}{!}{
\begin{tabular}{ccc||cc}
\specialrule{.2em}{.1em}{.1em}
\multirow{2}{*}{Scenario} & \multirow{2}{*}{Category} & \multirow{2}{*}{Model}  & \multicolumn{2}{c}{DIL} \\
 &  & & Acc. & MF1 \\
\specialrule{.1em}{.05em}{.05em}
\multirow{8}{*}{\begin{tabular}[c]{@{}c@{}}Non-continual\\      Learning\end{tabular}} & \multirow{2}{*}{BERT} & ONE & $\pm${0.0145} & $\pm${0.0300} \\
 &  & ONE+csc  & $\pm${0.0127}	& $\pm${0.0336}  \\
  & \multirow{2}{*}{BERT (Frozen)} & ONE & $\pm${0.0100} & $\pm${0.0024} \\
 &  & ONE+csc & $\pm${0.0140} & $\pm${0.0149} \\
 & \multirow{2}{*}{Adapter-BERT} & ONE & $\pm${0.0170} & $\pm${0.0379} \\
  &  & ONE+csc &  $\pm${0.0094} & 	$\pm${0.0327}\\
 & \multirow{2}{*}{W2V} & ONE & $\pm${0.0129}	& $\pm${0.0206}\\
 &  & ONE+csc &  $\pm${0.0092}	& $\pm${0.0079}\\
 \cline{1-5}
\multirow{40}{*}{\begin{tabular}[c]{@{}c@{}}Continual\\      Learning\end{tabular}} 
& \multirow{2}{*}{BERT} & NCL & $\pm${0.0137} & $\pm${0.0228}   \\
 &  & NCL+csc  &  $\pm${0.0266} & $\pm${0.0328} \\
& \multirow{2}{*}{BERT (Frozen)} & NCL & $\pm${0.0065} & $\pm${0.0110}  \\
 &  & NCL+csc  & $\pm${0.0085} & $\pm${0.0116}  \\
 & \multirow{2}{*}{Adapter-BERT} & NCL & $\pm${0.0102} & $\pm${0.0128} \\
  &  & NCL+csc  & $\pm${0.0102} & $\pm${0.0130}  \\
 & \multirow{2}{*}{W2V} & NCL & $\pm${0.0192} & $\pm${0.0230}\\
  &  & NCL+csc &  $\pm${0.0121} & $\pm${0.0108}\\
 \cline{2-5}
 & \multirow{10}{*}{\begin{tabular}[c]{@{}c@{}}BERT\\      (Frozen)\end{tabular}} & KAN & \multicolumn{2}{c}{---}\\
 &  & KAN+last & $\pm${0.0032}	& $\pm${0.0045}\\
 &  & KAN+ent &  $\pm${0.0056} & $\pm${0.0065} \\
 &  & SRK  & $\pm${0.0069} & $\pm${0.0099}\\
 &  & EWC  & $\pm${0.0094} & $\pm${0.0093}\\
 &  & UCL & $\pm${0.0084} & $\pm${0.0100}\\
 &  & OWM & $\pm${0.0092} & $\pm${0.0140}\\
 &  & DER++ & $\pm${0.0096} & $\pm${0.0120}\\
 &  & HAT  & \multicolumn{2}{c}{---}\\
 &  & HAT+last &  $\pm${0.0095} & $\pm${0.0098} \\  
 &  & HAT+ent &  $\pm${0.0029}	& $\pm${0.0097} \\
 \cline{2-5}
 & \multirow{6}{*}{Adapter-BERT} &  EWC & $\pm${0.0110} & $\pm${0.0110} \\
 &  & UCL & $\pm${0.0000} & $\pm${0.0000} \\
 &  & OWM & $\pm${0.0052} & $\pm${0.0109} \\
 &  & DER++ & $\pm${0.0137} & $\pm${0.0179}\\
 &  & HAT & \multicolumn{2}{c}{---} \\
 &  & HAT+last & $\pm${0.0038} & $\pm${0.0055}  \\  
 &  & HAT+ent & $\pm${0.0059} & $\pm${0.0106} \\
 \cline{2-5}
 & \multirow{10}{*}{W2V} & KAN & \multicolumn{2}{c}{---} \\	
 &  & KAN+last  & $\pm${0.0000} & $\pm${0.0000} \\  
 &  & KAN+ent  &  $\pm${0.0000} & $\pm${0.0000}  \\
 &  & SRK & $\pm${0.0000} & $\pm${0.0000}  \\
 &  & EWC &  $\pm${0.0059}	& $\pm${0.0076} \\
 &  & UCL & $\pm${0.0097} & $\pm${0.0128}\\
 &  & OWM & $\pm${0.0077} & $\pm${0.0081}\\
 &  & DER++ & $\pm${0.0075} & $\pm${0.0041}\\
 &  & HAT & \multicolumn{2}{c}{---}\\
 &  & HAT+last &  $\pm${0.0053} & $\pm${0.0082}\\  
 &  & HAT+ent  &  $\pm${0.0103} & $\pm${0.0199}\\
 \cline{2-5}
 & \multicolumn{2}{c||}{LAMOL} & $\pm${0.0027}	& $\pm${0.0062}\\
 \cline{2-5}
 & \multicolumn{2}{c||}{CLASSIC} & $\pm${0.0048} & $\pm${0.0101} \\
\specialrule{.1em}{.05em}{.05em}
\end{tabular}
}
\caption{Standard deviations. HAT and KAN have no results because they require task ids in testing, but in the DIL setting, no task ids are provided in testing.}
\vspace{-2mm}
\label{tab:std_results}
\end{table}

\begin{table}[]
\centering
\resizebox{\columnwidth}{!}{
\begin{tabular}{ccc||cccc}
\specialrule{.2em}{.1em}{.1em}
\multirow{2}{*}{Scenario} & \multirow{2}{*}{Category} & \multirow{2}{*}{Model} & \multicolumn{2}{c}{TIL} & \multicolumn{2}{c}{DIL} \\
 &  & & Acc. & MF1 & Acc. & MF1 \\
\specialrule{.1em}{.05em}{.05em}
\multirow{8}{*}{\begin{tabular}[c]{@{}c@{}}Non-continual\\      Learning\end{tabular}} 
& \multirow{2}{*}{BERT} & ONE & 0.8584 & 0.7635 & 0.8584 & 0.7635 \\
 &  & ONE+csc &  0.8353	& 0.7388 &  0.8353	& 0.7388  \\
  & \multirow{2}{*}{BERT (Frozen)} & ONE & 0.7814 & 0.5813 & 0.7814 & 0.5813 \\
 &  & ONE+csc &  0.8265 & 0.7232 &  0.8265 & 0.7232  \\
 & \multirow{2}{*}{Adapter-BERT} & ONE & 0.8596 & 0.7807 & 0.8596 & 0.7807 \\
  &  & ONE+csc &  0.8530 & 0.7516 &  0.8530 & 0.7516  \\
 & \multirow{2}{*}{W2V} & ONE & 0.7701 & 0.5189 & 0.7701 & 0.5189 \\
 &  & ONE+csc & 0.7761	& 0.5487  &  0.7761	& 0.5487  \\
 \cline{1-7}
\multirow{40}{*}{\begin{tabular}[c]{@{}c@{}}Continual\\      Learning\end{tabular}} 
& \multirow{2}{*}{BERT} & NCL & 0.4960 & 0.4308 & 0.8048 & 0.7085  \\
 &  & NCL+csc &  0.5939	& 0.3416 &  0.7727 & 0.5807 \\
& \multirow{2}{*}{BERT (Frozen)} & NCL & 0.8551 & 0.7664 & 0.8685 & 0.7873 \\
 &  & NCL+csc & 0.8783 & 0.8271 & 0.8693 & 0.7912 \\
 & \multirow{2}{*}{Adapter-BERT} & NCL & 0.5403 & 0.4481 & 0.8667 & 0.7804 \\
  &  & NCL+csc &  0.8630 & 0.8090 & 0.8809 & 0.7847  \\
 & \multirow{2}{*}{W2V} & NCL & 0.8269 & 0.7356 & 0.7736 & 0.6765 \\
  &  & NCL+csc &  0.8421 & 0.7418 & 0.8396 & 0.7509  \\

  \cline{2-7}
 & \multirow{10}{*}{\begin{tabular}[c]{@{}c@{}}BERT\\      (Frozen)\end{tabular}} & KAN & 0.8549 & 0.7738 & \multicolumn{2}{c}{---}\\
 &  & KAN+last & \multicolumn{2}{c}{---} & 0.8320 & 0.7352 \\
 &  & KAN+ent & \multicolumn{2}{c}{---} &  0.8278 & 0.7243 \\
 &  & SRK & 0.8476 & 0.7852 & 0.8391 & 0.7438 \\
 &  & EWC & 0.8637 & 0.7452 & 0.8660 & 0.7831 \\
 &  & UCL & 0.8389 & 0.7482 & 0.8538 & 0.7690 \\
 &  & OWM & 0.8702 & 0.7931 & 0.8611 & 0.7665 \\
 &  & DER++ & 0.8427 & 0.7508 &  0.8753 & 0.8009\\
 &  & HAT & 0.8674 & 0.7816 & \multicolumn{2}{c}{---}\\
 &  & HAT+last & \multicolumn{2}{c}{---} & 0.8473 & 0.7649 \\  
 &  & HAT+ent & \multicolumn{2}{c}{---} & 0.8418 & 0.7614  \\
 \cline{2-7}
 
  & \multirow{6}{*}{Adapter-BERT} &  EWC & 0.5630 & 0.4958 & 0.8805 & 0.7875 \\
 &  & UCL & 0.6446 & 0.3664 & 0.7123 & 0.3961 \\
 &  & OWM & 0.7299 & 0.6651 & 0.8766 & 0.7882 \\
 &  & DER++ & 0.4763 & 0.3554& 0.8859 & 0.7985 \\
 &  & HAT & 0.8614 & 0.7852 & \multicolumn{2}{c}{---} \\
 &  & HAT+last & \multicolumn{2}{c}{---} & 0.8823 & 0.7919 \\  &  & HAT+ent & \multicolumn{2}{c}{---} & 0.8854 & 0.8245 \\
 \cline{2-7}
 
  & \multirow{10}{*}{W2V} & KAN & 0.7206 & 0.4001 & \multicolumn{2}{c}{---} \\	
   &  & KAN+last & \multicolumn{2}{c}{---} &  0.7123 & 0.3961 \\  
 &  & KAN+ent & \multicolumn{2}{c}{---} & 0.7123 & 0.3961  \\
 &  & SRK & 0.7101 & 0.3963 & 0.7123 & 0.3961\\
 &  & EWC & 0.8416 & 0.7229 & 0.7586 & 0.6545\\
 &  & UCL & 0.8441 & 0.7599 & 0.8187 & 0.6965 \\
 &  & OWM & 0.8270 & 0.7118 & 0.8256 & 0.7253\\
 &  & DER++ & 0.8327 & 0.6993 & 0.8459 & 0.7722\\
 &  & HAT & 0.8083 & 0.6363 & \multicolumn{2}{c}{---}\\
 &  & HAT+last & \multicolumn{2}{c}{---} &  0.7599 & 0.5849  \\  
 &  & HAT+ent & \multicolumn{2}{c}{---} &  0.7605 & 0.5349  \\
  \cline{2-7}
 & \multicolumn{2}{c||}{LAMOL} & 0.8891 & 0.8059 & 0.8891 & 0.8059 \\
  \cline{2-7}
 & \multicolumn{2}{c||}{\textbf{CLASSIC (forward)}} & \textbf{0.8897} & \textbf{0.8338} & \textbf{0.8886} & \textbf{0.8365} \\
 & \multicolumn{2}{c||}{\textbf{CLASSIC}} & \textbf{0.8942} & \textbf{0.8393} & \textbf{0.9022} & \textbf{0.8512} \\
\specialrule{.1em}{.05em}{.05em}
\end{tabular}
}
\caption{Accuracy (Acc.) and Macro-F1 (MF1) averaged over 5 random sequences of 19 tasks. 
\vspace{-2mm}
} 
\label{tab:all_results}
\end{table}

\subsection{Standard Deviations}
Table \ref{tab:std_results} reports the standard deviation of CLASSIC and the considered baselines over 5 runs with random seeds using one random task sequence. We can see the results are stable. 

\subsection{CLASSIC in TIL Scenario}

Table \ref{tab:all_results} shows that our proposed method CLASSIC can be adapted for the \textit{Task Incremental Learning} (TIL) setting of continual learning, which requires the task ids during testing, but our DIL setting does not require. We can observe that CLASSIC in the TIL setting also outperforms existing TIL baselines.

\subsection{Execution Time and Number of Parameters}

Table \ref{tab:parameter_time} reports the number of parameters (regardless of trainable or non-trainable) and the training execution times of different models. The execution time is computed as the average training time \textit{per} task. 
Our experiments were run on GeForce GTX 2080 Ti with 11G GPU memory.

\begin{table}[t]
\centering
\resizebox{\columnwidth}{!}{
\begin{tabular}{ccc||cc}
\specialrule{.2em}{.1em}{.1em}
Scenarios & Category & Model & \#parameters (M) & Running time (s) \\
 \specialrule{.1em}{.05em}{.05em}
\multirow{4}{*}{\begin{tabular}[c]{@{}c@{}}Non-continual   \\      Learning \end{tabular}} & \multicolumn{1}{l}{BERT} & ONE & 109.5 & 600.0 \\
  & BERT (Frozen) & ONE & 110.4 & 500.0 \\
 & Adapter-BERT & ONE & 183.3 & 684.0  \\
 & W2V & ONE & 6.7 & 189.0  \\
 \cline{1-5}
\multirow{21}{*}{\begin{tabular}[c]{@{}c@{}}Continual \\      Learning\end{tabular}} & BERT & NCL & 109.5 & 600.0 \\
 & BERT (Frozen) & NCL & 110.4 & 500.0  \\
 & Adapter-BERT & NCL & 183.3 & 684.0  \\
 & W2V & NCL & 6.7 & 189.0  \\ \cline{2-5}
 & \multirow{6}{*}{BERT   (frozen)} & KAN & 116.6 & 550.0 \\
 &  & SRK & 117.8 & 600.0  \\
 &  & EWC & 110.4 & 580.0  \\
 &  & UCL & 110.4 & 638.0  \\
 &  & OWM & 110.6 & 635.0  \\
 &  & DER++ & 115.0 & 650.0 \\
 &  & HAT & 111.3 & 610.0 \\ \cline{2-5}

 & \multirow{4}{*}{Adapter-BERT} & EWC & 183.3 & 840.0  \\
 &  & UCL & 183.4 & 870.0  \\
 &  & OWM & 184.4 & 780.0 \\
  &  & DER++ & 184.0 & 880.0 \\
 &  & HAT & 185.2 & 840.0 \\  \cline{2-5}

 & \multirow{6}{*}{W2V} & KAN & 7.0 & 150.0  \\
 &  & SRK & 7.2 & 160.0  \\
 &  & EWC & 6.2 & 162.0  \\
 &  & UCL & 6.2 & 135.0  \\
 &  & OWM & 6.4 & 125.0 \\
 &  & DER++ & 6.8 & 135.0 \\
 &  & HAT & 6.4 & 180.0  \\  \cline{2-5}
 & \multicolumn{2}{c||}{LAMOL} & 124.4 & 650.0 \\ \cline{2-5}
 & \multicolumn{2}{c||}{CLASSIC} &  185.2 & 900.0	 \\

\specialrule{.1em}{.05em}{.05em}

\end{tabular}
}
\caption{Network size (number of parameters, regardless of trainable or non-trainable) and {\color{black}average training execution time per task} of each model in seconds. 
\vspace{-4mm}
} 
\label{tab:parameter_time}
\end{table}

\subsection{Hyper-parameters and Validation Results}

Sec.~4.3 in the main paper reported the best hyper-parameters. Regarding hyper-parameter search, we performed grid search on the \textit{temperature} parameter $\tau$ within \{0.03, 0.5, 0.8, 1\}, \textit{batch size} within \{32, 64, 128\}, and $s_{max}$ within \{140, 200, 300, 400\}. We also experimented with whether to apply the $l_2$ normalization before contrast and whether to use the \textit{logits} or the \textit{second last} layer to do the contrast. {\color{black}We did not save the validation results but the reported test results in the paper are given by the parameters with the best validation performance.} We encourage the reviewers and interested readers to play with the submitted code.



\end{document}